\documentclass[]{cap2024}

\usepackage{multicol} 
\usepackage{fancyvrb}
\usepackage{amsfonts} 
\usepackage{amssymb}
\usepackage{amsmath}
\usepackage{booktabs} 
\usepackage{cleveref}
\usepackage{graphicx}
\usepackage{multirow}
\usepackage{subcaption} 
\usepackage{float}
\usepackage{enumitem}

\usepackage{hyperref}   
\hypersetup{
  colorlinks   = true, 
  urlcolor     = purple, 
  linkcolor    = red, 
  citecolor    = blue 
}

\usepackage{todonotes}

\title{\textbf{RelevAI-Reviewer}: A Benchmark on AI Reviewers for Survey Paper Relevance}

\author[1]{Paulo Henrique Couto\thanks{These authors contributed equally to this work.}\thanks{Corresponding author: paulo.couto-de-resende-silva@universite-paris-saclay.fr}}
\author[1]{Quang Phuoc Ho\textsuperscript{*}}
\author[1]{Nageeta Kumari\textsuperscript{*}}
\author[1]{Benedictus Kent Rachmat}
\author[1]{Thanh Gia Hieu Khuong}
\author[2]{Ihsan Ullah}
\author[1]{Lisheng Sun-Hosoya}
\affil[1]{Université Paris-Saclay, France} 
\affil[2]{ChaLearn, USA}
\affil[ ]{\url{https://github.com/paulohenriquecrs/RelevAI-Reviewer}}
\affil[ ]{Manuscript of paper accepted to \textit{Conférence sur l'Apprentissage automatique} (CAp) 2024}

\begin{document}
\maketitle
\begin{abstract}
  Recent advancements in Artificial Intelligence (AI), particularly the widespread adoption of Large Language Models (LLMs), have significantly enhanced text analysis capabilities. This technological evolution offers considerable promise for automating the review of scientific papers, a task traditionally managed through peer review by fellow researchers. Despite its critical role in maintaining research quality, the conventional peer-review process is often slow and subject to biases, potentially impeding the swift propagation of scientific knowledge. In this paper, we propose \textbf{RelevAI-Reviewer}, an automatic system that conceptualizes the task of survey paper review as a classification problem, aimed at assessing the relevance of a paper in relation to a specified prompt, analogous to a ``call for papers". To address this, we introduce a novel dataset comprised of $25\,164$ instances. Each instance contains one prompt and four candidate papers, each varying in relevance to the prompt. The objective is to develop a machine learning (ML) model capable of determining the relevance of each paper and identifying the most pertinent one. We explore various baseline approaches, including traditional ML classifiers like Support Vector Machine (SVM) and advanced language models such as BERT. Preliminary findings indicate that the BERT-based end-to-end classifier surpasses other conventional ML methods in performance. We present this problem as a public challenge to foster engagement and interest in this area of research.
\medskip

\end{abstract}

\section{Introduction}
\label{seclentete}
The advent of artificial intelligence (AI) has revolutionized diverse domains, promising to enhance efficiency and reduce human bias in tasks that require nuanced understanding. One such domain ripe for innovation is the scientific peer-review process. Traditionally, this process has relied heavily on human expertise, which, while invaluable, presents several inherent challenges, as discussed by \cite{nihar2024}. These challenges include inconsistencies in reviewer judgments, potential human biases or conflicts of interest, and the extensive time commitment required for thorough evaluations. In this context, AI has the potential to provide a more standardized, impartial, and efficient review process, thereby enhancing the overall quality, efficiency, and fairness of scientific assessments.

In this paper, we introduce an AI-powered reviewer: \textbf{RelevAI-Reviewer}, designed to address the mentioned challenges in the traditional reviewing process, emphasizing the critical attributes of relevance in evaluating scientific manuscripts, particularly survey papers. This AI-assisted reviewer operates by assessing the relevance of four different candidate survey papers in response to a specific ``call for paper" prompt, aiming to identify the paper that is most relevant to the prompt. Additionally, this reviewer can also function as a classifier to tackle a four-class classification problem with a prompt-paper pair. It predicts the relevance of the paper to the prompt, assigning it to one of four possible relevance classes—\{0, 1, 2, 3\}—ranging from the least relevant (lowest) to the most relevant (highest).

\subsection{Problem Formulation}
\cite{rachmat2023auto} in their work of organizing the pioneering Auto-Survey challenge, identified five key metrics for evaluating an AI-reviewer's efficacy: \textbf{``Relevance"}, \textbf{``Contribution"}, \textbf{``Soundness"}, \textbf{``Clarity"}, and \textbf{``Responsibility"}. In particular, Relevance assesses whether the content of a paper aligns with the specified prompt, similar to a ``Call for Papers" in the scientific publication cycle. As this particular criterion is among the most important aspects in the peer review process, we decided to focus solely on it in this study and reserve the other criteria for future work.

The prompts are created through a reverse-engineering process, taking the form ``\textit{Write a systematic survey or overview about [paper title + paper abstract].}". \cite{rachmat2023auto} evaluate relevance by using cosine similarity between the embeddings of the prompt and the paper. Building on this foundation, we propose enhancing this criterion's evaluation by developing an AI model, trained to assess the relevance of candidate papers with the prompt, where the input comprises the pairs of \textit{prompt + candidate paper}, and the output is the predicted relevance level/class. Our goal is to formulate the problem as a standard machine learning task, specifically a four-class classification, to facilitate a more comprehensive training phase compared to the limited opportunities in pure prompt engineering. This approach maintains reasonable computational costs, in contrast to methods such as fine-tuning LLMs. Additionally, by opening this challenge as a public benchmark (\autoref{sec: The RelevAI-Reviewer Benchmark}), this familiar framework can encourage greater engagement and participation.

To achieve this, we have compiled a dataset from \href{https://www.semanticscholar.org/}{Semantic Scholar} [\cite{jones2015artificial}] survey papers\footnote{The reason for choosing survey papers for this study is that they typically contain fewer tables, theories, figures, or plots, which are challenging for current language models to parse effectively. This allows us to focus primarily on the text.}, constructing reverse-engineered prompts and pairing them with four candidate papers of varying relevance. The ground truth of their relevance is represented as an integer with four possible values, each denoting the relevance class of the corresponding candidate: \textit{\{0: least relevant, 1: second least relevant, 2: second most relevant, 3: most relevant\}}. According to the models used in the methodologies, these labels can be encoded in different ways, like one-hot encoding or thermometer encoding (see \autoref{label-ecoding}). The whole process of dataset creation is described in \autoref{Dataset-Creation}.

To solve this problem, we explore various approaches, including traditional machine learning (ML) classifiers augmented by sentence transformers for input vectorization, as well as end-to-end large language models (LLMs), fine-tuned on previous dataset. A detailed exposition of these methodologies will be provided in \autoref{Methodologies}.

\subsection{Related Work}
Prior research has explored automating aspects of the peer review process using Natural Language Processing (NLP) techniques. \cite{Ramachandran2017} investigated the use of NLP allied with machine learning techniques to measure the quality of peer reviews produced by fellow researchers. \cite{Chitra2016} applied machine learning algorithms such as Support Vector Machines (SVM) to improve paraphrased plagiarism detection, while \cite{Guo2023} assessed LLMs performance in identifying relevant titles and abstracts from clinical review datasets, comparing it to human reviewers and highlighting its ability to provide reasoning for decisions and correct initial classifications when questioned. Despite the progress, challenges persist in integrating AI assistance into peer review workflows with fairness and reliability, as shown by \cite{Zhang2022}. This underscores the complexity of automating scholarly evaluation processes, which demand not only precision but also a nuanced understanding of academic content and ethics.

Parallel to these investigations, research work in exploring different computational techniques to improve the quality of peer reviews has considerably increased over the past years. For example, \cite{Saveski2023} investigated text similarity techniques to optimize the assignment of reviewers to submissions across various computer science disciplines. Furthermore, \cite{rachmat2023auto} delved into utilizing cosine similarity to assess the relevance between the content of candidate papers and the given prompt. Building upon these efforts, our study seeks to advance the field by developing and training a classifier on a tailored dataset, using prompt engineering techniques and Large Language Models (LLMs). Our approach sets the cosine similarity method as a baseline.

\section{Creation of the RelevAI-Reviewer Dataset}
\label{Dataset-Creation}
In this section, we detail the creation of the RelevAI-Reviewer Dataset, which comprises prompts and candidate papers derived from a wide range of academic fields, including Chemistry, Mathematics, and Social Sciences. The extensive list of all 22 scrapped subjects is available in \autoref{subject-scrapped}.

\subsection{Data Collection and Categorization}
\label{paper-scrapping}
Our primary objective was to evaluate paper relevance through a structured dataset comprising four categories of survey papers, each reflecting a different level of relevance to a command prompt. To manage computational and storage demands efficiently, each paper within the dataset is represented by its title, abstract, and a synthetic Related Work section.

Initially, we gathered approximately $30\,000$ papers, which were refined to $25\,164$ high-quality entries after a rigorous quality control process. This selection was further categorized into four relevance levels—most relevant, second most relevant, second least relevant, and least relevant—based on their relation to a reverse-engineered command prompt. The process for categorizing papers is as follows:
\begin{itemize}[leftmargin=*, itemsep=0pt, parsep=0pt]
  \item \textbf{prompt: } Reverse engineered command prompt (see \autoref{prompt-generation}) generated using the most\_relevant paper as reference;
  \item \textbf{most\_relevant: } Papers directly used to generate the command prompt, possessing the highest relevance;
  \item \textbf{second\_most\_relevant: } Papers cited by the most relevant paper, maintaining strong relevance;
  \item \textbf{second\_least\_relevant: } Randomly selected papers from the same field as the most relevant paper, showing moderate relevance;
  \item \textbf{least\_relevant: } Papers from different fields than the most relevant paper, exhibiting the least relevance.
\end{itemize}
An example instance of the dataset can be found in \autoref{appendix-dataset-example}.

\subsection{Reverse Prompt Engineering}
\label{prompt-generation}
To generate command prompts, we utilized the most relevant papers' \{{\it{title}}\} and \{{\it{abstract}}\} as input. Using the \cite{openai_api} API with the gpt-3.5-turbo model, we reverse-engineered prompts by submitting the titles and abstracts of these papers. The model was tasked with creating instructions that begin with ``You have a critical thinking mind. Generate a prompt to be input into large language models starting with `Write a systematic survey or overview about'. Limit yourself to 2 sentences. Title : \{{\it{title}}\} Abstract: \{{\it{abstract}}\}". To balance creativity with consistency in the responses, we set the model's temperature to 0.1.

\subsection{Artificial Related Work Section Generation and Quality Control}
\label{artificial-related-work-quality-control}
Given the Semantic Scholar API's limitations in accessing specific paper sections, we generated artificial Related Work sections for our dataset. This was necessary when direct access to a paper's Related Work was unavailable or inadequately parsed. Utilizing OpenAI's ChatGPT with the gpt-3.5-turbo model, we created these sections by referencing the title and abstract of three papers cited in the original work, ensuring the generated content maintained relevance and continuity.

To uphold quality standards, we selected cited papers based on their availability and citation count within Semantic Scholar, discarding any paper from the dataset that failed these quality checks. This approach, despite its constraints, allowed us to efficiently compile a dataset of $25\,164$ instances, each comprising a command prompt and four categorized papers (\textit{title}, \textit{abstract}, and \textit{artificial Related Work section}), refined from an initial collection of around $30\,000$ papers.

\subsection{Final Dataset}
We finalized the dataset with $25\,164$ instances, each comprising a command prompt and four papers categorized by relevance. For supervised learning applications, we segmented each instance into four separate entries, pairing each prompt with a single paper and its relevance label. This reformatting yields a total of $100\,656$ data points, organized into three columns: prompt, paper candidate, and relevance category.

\begin{figure}[htbp]
  \centering
  \includegraphics[width=0.85\textwidth]{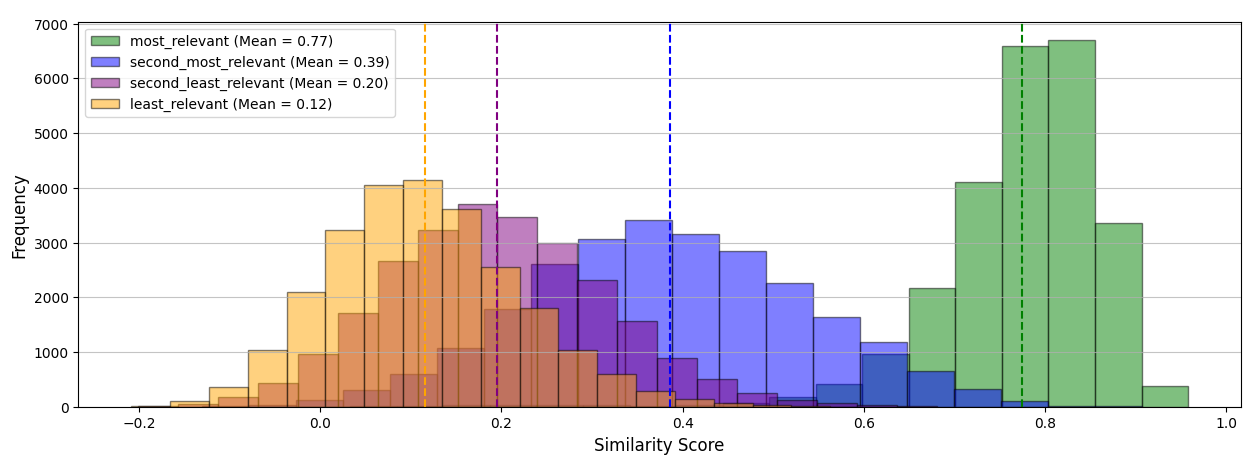}
  \caption{Distribution of cosine similarities between prompts and papers across four relevance categories using vectorized embeddings. Histograms were generated from similarity scores with the number of bins set to 20. Dashed lines represent the mean similarity score for each category.\label{fig:cosine_similarities_dist}}
\end{figure}

Analysis of the cosine similarity distributions across the paper-prompt pairs in \autoref{fig:cosine_similarities_dist} reveals clear distinctions in relevance. The most relevant papers exhibit the highest mean similarity, distinctively setting them apart from the less relevant categories. This separation is notably pronounced between the most relevant and the second most relevant categories, illustrating minimal overlap in their distribution ranges. Such distinctions underscore the effectiveness of our command prompt generation method, particularly how the most relevant papers align closely with the generated prompts. The variance among categories, from most to least relevant, reflects the nuanced degrees of relevance that our dataset captures, offering a robust foundation for evaluating and enhancing text similarity metrics in supervised learning models.

\section{RelevAI-Reviewer Methodology}
Using the RelevAI-Reviewer dataset described in the previous section, our objective is to develop an AI model trained with this dataset, to assess the relevance of a candidate paper and its generated ``call for paper" command prompt. In this section, we detail our approaches to label encoding, evaluation metrics, and the classification models employed in our RelevAI-Reviewer baselines.

\label{Methodologies}
\subsection{Label Encoding}
\label{label-ecoding}
Label encoding is a crucial step in the preprocessing phase, especially when dealing with a categorical target, as is the case here. The choice of encoding method can influence the information representation in the target, subsequently affecting the performance of the entire ML pipeline. In our study, we explored two distinct encoding approaches and evaluated their impact on the overall performance.
\begin{itemize}
    \item One-Hot Encoding: each relevance rank is represented as a binary 4-dimensional vector, where each dimension corresponds to a unique category and is either 1 or 0 to indicate the presence or absence of that category. To determine the read-out value, we extract the index of the largest logits value in the output vector. One limitation of this encoding approach is the possibility of two papers with the same prompt sharing the same relevance rank, as we take the index of the largest value in the output vectors and the vectors are independent of one another.
    \label{label-ecoding-onehot}
    \item Thermometer Encoding: each bit of the label vector differentiates between specific classes. Specifically, the first bit distinguishes class `0' from the rest; the second bit distinguishes between classes `0' and `1' from classes `2' and `3'; and the third bit distinguishes class `3' from the rest.\\
    We use binary cross-entropy loss and subsequently apply the softmax function to obtain the output vector. The output vectors of the four papers sharing the same prompt are then combined, and the sum of the four vectors is ranked. Relevance ranks are assigned as follows: `3' for the highest sum, `2' for the second highest, `1' for the third highest, and `0' for the lowest sum, as illustrated in \autoref{tab:combined_encoding_comparison}.
\end{itemize}

\begin{table}[H]
\centering
\begin{minipage}[t]{0.3\textwidth}
    \centering
    \scriptsize
    \begin{tabular}{|p{1.5cm}|p{1.5cm}|p{2cm}|}
    \hline
    \textbf{Relevance Rank} & \textbf{One-hot Vector} & \textbf{Thermometer Vector} \\
    \vspace*{\fill} & \vspace*{\fill} & \vspace*{\fill} \\
    \hline
    3 & [0, 0, 0, 1] & [1, 1, 1] \\
    2 & [0, 0, 1, 0] & [1, 1, 0] \\
    1 & [0, 1, 0, 0] & [1, 0, 0] \\
    0 & [1, 0, 0, 0] & [0, 0, 0] \\
    \hline
    \end{tabular}
\end{minipage}
\hfill
\begin{minipage}[t]{0.62\textwidth}
    \centering
    \scriptsize
    \begin{tabular}{|c|c|c|}
    \hline
    \textbf{Encoding} & \textbf{Output} & \textbf{Relevant Rank} \\
    \hline
    \multirow{2}{*}{One-hot encoding}  & \textbf{Paper 1:} [0.9, -7.1, 5, 2] & 2 \\
    & \textbf{Paper 2:} [-5.3, 2.5, 4, 1] & 2 \\
    \hline
    \multirow{4}{*}{Thermometer encoding} & \textbf{Paper 1:} [0.91, 0.82, 0.17], sum=1.9 & 2 \\
    & \textbf{Paper 2:} [0.94, 0.1, 0.3], sum=1.34 & 1 \\
    & \textbf{Paper 3:} [0.23, 0.02, 0.1], sum=0.35 & 0 \\
    & \textbf{Paper 4:} [0.90, 0.89, 0.96], sum=2.75 & 3 \\
    \hline
    \end{tabular}
\end{minipage}
\label{tab:label_encoding_comparision}
\caption{Comparison of relevance rank encoding using one-hot and thermometer encoding methods (left), and their impact on output determination (right).}
\label{tab:combined_encoding_comparison}
\end{table}

\subsection{Evaluation}
We evaluate the performance of models by computing the mean and standard error of both Kendall’s Tau score and F1-score for each class. Kendall’s Tau score is a measure of the correlation between two rankings and is often used to assess the agreement between two sets of ordered data [\cite{Kendall1938}]. The Kendall’s Tau score, denoted as $K$, is calculated based on the number of concordant (C) and discordant (D) pairs in the data: $K = (C - D) / (C + D)$. This metric provides an assessment of the model's overall ranking accuracy across all classes.

The F1-score is a metric representing the harmonic mean of precision and recall \cite{Powers2021}. Given a binary classification task, the F1-score is computed as:
$F1 = 2 * Precision * Recall / (Precision + Recall)$.
This metric provides the evaluation of a model's performance on a particular class.

\subsection{Classification}
In our analysis using classification models, our primary focus is on comparing traditional machine learning models against a more expensive model, namely BERT.
\subsubsection{Sentence Transform + Scikit-learn Classifiers}
\label{subsubsec: SentenceTransformer +  ML Classifiers}
SentenceTransformers\footnote{https://huggingface.co/sentence-transformers} is a Python framework for state-of-the-art sentence, text, and image embeddings. We used classifiers with SentenceTransformer's [\cite{reimers2019sentence}] paraphrase-MiniLM-L6-v2\footnote{https://huggingface.co/sentence-transformers/paraphrase-MiniLM-L6-v2} model from Hugging Face to generate dense vectors from our text input. The reason for using SentenceTransformer lies in the fact that our goal was to determine the relevance of the paper to a given prompt, and for this, a higher level of context understanding was required.  
The paraphrase-MiniLM-L6-v2 SentenceTransformer maps sentences or paragraphs to a 384-dimensional vector, making it suitable for tasks such as semantic search and clustering. 
We used these embeddings with multiple classification models from scikit-learn [\cite{scikit-learn}].

\subsubsection{BERT}
\label{BERT-intro}
BERT (Bidirectional Encoder Representations from Transformers) is a pre-trained transformer-based model introduced by \cite{Devlin2018} at Google. It excels in natural language processing tasks by bidirectionally capturing deep contextual relationships within text data, leading to state-of-the-art performance across various NLP applications. For example, \cite{koroteev2021bert} showed application and adaptation of the BERT model to problems such as question answering, text classification, and named entity recognition.
In our experiment, we employed the BERT sequence classification approach to learn the relevance ranks. Particularly, we leveraged the bert-based-uncase\footnote{https://huggingface.co/google-bert/bert-base-uncased} pre-trained model, which consists of 12 layers, a hidden size of 768, and 12 self-attention heads, totaling 110 million parameters, for tokenization and training.

\section{Experimental Results}
\label{section:experimental-results}
In this section, we present the results of our experiments. The dataset was divided using an 80\%--20\% train-test split. To assess data efficiency and study the relation between performance potential and dataset size, we developed some experiments with increasing sizes for a sub-sample of the train set, as detailed in \autoref{appendix-f1-train-sizes} and \autoref{appendix-Bert-Results}.

\subsection{Scikit-Learn Classifiers}
We used four classifiers available in scikit-learn, Support Vector Machine (SVM [\cite{cortes1995support}]), K Nearest Neighbors (KNN [\cite{cover1967nearest}]), Random Forest (RF [\cite{ho1995random}]), and Logistic Regression [\cite{cox1958regression}] via SGD, and conducted hyperparameter tunning using RandomizedSearchCV. The details and best hyperparameters are given in \autoref{appendix-hyperparameter-tuning}.
The models input was the vectorized representation of the \textit{prompt + candidate paper} pair, returned by the sentence transformer as described in Section \ref{subsubsec: SentenceTransformer +  ML Classifiers}, with a label indicating the relevance category as output.

\autoref{tab:performance} presents the results for all four models, where SVC gives the highest Kendall's Tau (0.76) and F1-score out of the four scikit-learn classifiers. Based on these results, we selected SVC to further explore its data efficiency by varying the sizes of the training set while keeping the test sample size constant across all experiments. These results are detailed in \autoref{appendix-f1-train-sizes}. We found that SVC demonstrates robust performance even with limited training data, particularly in identifying the most relevant candidates. This effectiveness may be attributed to its capability to identify the optimal separating hyperplane with the maximum margin, coupled with its proficiency in managing high-dimensional feature spaces.

\subsection{BERT}

\label{BERT}
We conducted two experiments using one-hot and thermometer encoded labels. In each experiment, we trained the bert-base-uncased model, as discussed in  Section \ref{BERT-intro}, with various training sets. We employ a batch size of 4 and utilize the Adam optimizer, a popular choice in deep learning optimization algorithms [\cite{Kingma2014}]. For the training setup, we used the Kaggle\footnote{\url{https://www.kaggle.com/}} platform, which provided P100 GPUs with 16GB RAM. The training time for one epoch with a training set consisting of $80\,524$ examples was approximately 1 hour and 15 minutes.

The result presented in \autoref{tab:performance} shows the performance when training with the full train set, which contains $80\,524$ examples, using two approaches: one-hot (1) and thermometer (2) encoding (see \autoref{label-ecoding}). For (1), the mean of Kendall's Tau score is 0.893 with a standard error of 0.002. For (2), the score is improved by approximately 3\%, with a similar standard error. This improvement can be explained by the fact that (2) avoids repeating classes for papers from the same prompt. In contrast, the one-hot encoding approach can classify two papers from the same prompt into the same category. While this is not an issue for classification tasks, it poses a problem for ranking tasks.

We also further investigated data efficiency in the computationally intensive BERT models. Despite the overall performance improvement with increasing training set sizes, we identify a noteworthy pattern at approximately $5\,032$ examples—less than 10\% of the total training size. At this point, BERT models, regardless of using one-hot or thermometer encoding, start to approach a performance plateau. This trend is clear in our primary metrics: Kendall's Tau for overall ranking accuracy and the F1-score for the most relevant class, which is of primary interest. This observation indicates that it is possible to achieve significant savings on training resources by reducing the number of training examples, without compromising performance. For a detailed discussion, refer to \autoref{appendix-Bert-Results}.

\subsection{Comparison of Performances Among Different Baselines}
\label{comparison-performances}

\autoref{tab:performance} presents an analysis of Kendall's Tau scores for all baseline models. In our simplest approach, we perform a classification based on fixed thresholds for the cosine similarity between the vector representations of prompts and papers. In this approach, we use the train set to search in a grid of possible values for the three thresholds in order to find the ones that, when used to classify the instances of the train data into the four relevance categories, provide the highest Kendall's Tau. More details on this methodology can be found on \autoref{apx-cosine-thresholds}.

The performance of each model, including Kendall's Tau and F1-scores along with their respective bootstrap standard errors, is presented in \autoref{tab:performance}. BERT models significantly surpass the SVC and cosine similarity baseline, demonstrating their robustness in capturing semantic relationships. Among the experiments, BERT with thermometer encoding achieves the highest Kendalls’ Tau score, suggesting that this encoding method enhances the model’s performance when predicting rankings. The standard error for BERT models, which indicates variability in the bootstrap test results, is the smallest for the BERT\textsubscript{thermometer}, demonstrating consistent performance across multiple test samples.

\begin{table}[htbp]
    \centering
    \footnotesize
    \caption{Performance comparison of different approaches}
    \label{tab:performance}
    \noindent\hspace*{-2em}
    \begin{tabular}{lcccccc}
        \toprule
        Model & Kendall’s Tau score && F1-score \\
        \midrule
        & & Most & Second most & Second least & Least \\
        \midrule
        Cosine Similarity Baseline & 0.774 $\pm$ .003 & 0.943 $\pm$ .002 & 0.091 $\pm$ .005 & 0.250 $\pm$ .006 & 0.634 $\pm$ .004\\
        Support Vector Machine & 0.761 $\pm$ .003 & 0.947 $\pm$ .002 & 0.691  $\pm$ .005 & 0.472  $\pm$ .006 & 0.619  $\pm$ .005 \\
        K Nearest Neighbors & 0.726  $\pm$ .003 & 0.942  $\pm$ .002 & 0.645 $\pm$ .005 & 0.418  $\pm$ .006 & 0.509  $\pm$ .006 \\
        Random Forest & 0.747  $\pm$ .003 & 0.939  $\pm$ .002 & 0.695  $\pm$ .005 & 0.360  $\pm$ .006 & 0.629  $\pm$ .005 \\
        Logistic Regression & 0.738  $\pm$ .003 & 0.887  $\pm$ .003 & 0.541  $\pm$ .006 & 0.264  $\pm$ .006 & 0.629  $\pm$ .004 \\
        BERT (one-hot) & 0.893 $\pm$ .002 & 0.990 $\pm$ .001 & 0.941 $\pm$ .002 & 0.733 $\pm$ .005 & 0.739 $\pm$ .005 \\
        \textbf{BERT (thermometer)} & \textbf{0.928 $\pm$ .002} & \textbf{0.994 $\pm$ .001} & \textbf{0.965 $\pm$ .002} & \textbf{0.807 $\pm$ .004} & \textbf{0.827 $\pm$ .004} \\
        \bottomrule
    \end{tabular}
\end{table}

\section{The RelevAI-Reviewer Benchmark}
\label{sec: The RelevAI-Reviewer Benchmark}
We believe the AI-assisted reviewer holds considerable potential for practical implementation. The work presented in this paper serves as an initial investigation. To engage the AI community and foster further research in this domain, we are introducing this problem as an open benchmark on the Codabench platform \cite{pavao2023codalab}, where we provide access to the Relevance-AI dataset and our baseline models' code. All submitted papers, training, and testing will be processed on our server to ensure data protection, using locally installed models, including the pre-trained BERT. We invite interested parties to participate. For more details and to join the challenge, please visit the benchmark website: \href{https://www.codabench.org/competitions/1946/}{https://www.codabench.org/competitions/1946/}

\section{Conclusion and Future work}
This study has introduced and examined the performance of RelevAI-Reviewer, an AI-powered tool designed to enhance the scientific peer-review process by assessing the relevance of survey papers to specific prompts. Our findings underscore the potential of incorporating AI to augment the traditional, manual peer-review process, offering a promising avenue toward reducing human bias and time constraints. The core contribution of this work is the development of an AI model capable of effectively ranking scientific manuscripts in terms of relevance, leveraging a novel dataset specifically curated for this purpose. Our experiments reveal that BERT models, particularly when coupled with thermometer encoding, significantly outperform traditional machine learning approaches in this context. 

Despite the promising results, this study is not without limitations. The scope was confined to survey papers, which represent a fraction of the broader scientific corpus. Also, the construction of the dataset, which emphasizes titles, abstracts, and artificially generated related work sections, may not capture the full depth of papers as would a comprehensive review of the entire text. Furthermore, several assumptions in the current methodology could introduce limitations and potential biases: using prompts generated based on available papers may introduce biases into the dataset; relying on human-cited papers as ground truth can be questionable, as some uncited papers might actually be more relevant than the cited ones; and the formulation into a 4-class classification may have limitations, as two papers can share the same relevance level. Addressing these limitations will be the focus of our future research. In particular, we will focus on expanding the dataset to include a wider variety of document types and disciplines, exploring the integration of full-text analysis, and experimenting with new similarity metrics. Additionally, new experiments could be conducted using different LLMs to compare performance, which would require minimal adjustments at the data level. Finally, it would be beneficial to expand this reviewer by incorporating approaches that provide reasoning alongside the score. This would justify the predicted relevance class to the author and offer guidance on how to improve relevance. These features present promising avenues for future exploration.

\section{Acknowledgements}
This work was supported by ChaLearn and conducted as part of the ``Creation of an AI Challenge" master's class of Universit\'e Paris-Saclay in the 2024 academic year. 
We express our deepest gratitude to Isabelle Guyon for her pivotal role in establishing this class and the invaluable suggestions and feedback, enriching all aspects of this project. We thank the administrators of Codabench for providing the benchmark platform and assisting in setting up the benchmark. Supported by ANR Chair of Artificial Intelligence HUMANIA ANR-19-CHIA-0022 and TAILOR EU Horizon 2020 grant 952215.

\bibliographystyle{apalike}
\bibliography{cap2024}

\newpage 
\appendix 

\section{Relevance-AI Dataset Example}
\label{appendix-dataset-example}

Here we demonstrate one example instance of the dataset to illustrate the difference in the subjects of the papers from each relevance category considering the command prompt:

\textit{``Write a systematic survey or overview about the prevalence and determinants of beliefs in conspiracy theories related to the COVID-19 pandemic in a nationally representative sample of internet users. The survey should include an analysis of the association of support for conspiracy theories with sociodemographic variables, health literacy, and eHealth literacy."}

\medskip

\renewcommand{\arraystretch}{1.2} 
\begin{table}[htbp]
  \centering  
  \begin{tabular}{p{0.25\textwidth} p{0.75\textwidth}}
    \toprule
    \textbf{Relevance Categories} & \textbf{Instance Example} \\ \midrule
    most\_relevant & \textbf{Title:} The Determinants of Conspiracy Beliefs Related to the COVID-19 Pandemic in a Nationally Representative Sample of Internet Users \\ & \textbf{Abstract:} An overwhelming flood of misinformation is accompanying the pandemic of COVID-19. Fake news and conspiracy theories are so prevalent that the World Health Organization started as early as February 2020 [...] \\ & \textbf{Related Work:} The COVID-19 pandemic has given rise to a concerning prevalence of misinformation and conspiracy theories, leading to the term ``infodemic" coined by the World Health Organization [...] \\ 
    \hline
    second\_most\_relevant & \textbf{Title:} The science of fake news \\ & \textbf{Abstract:} Addressing fake news requires a multidisciplinary effort. The rise of fake news highlights the erosion of long-standing institutional bulwarks against misinformation in the internet age. Concern over the problem [...] \\ & \textbf{Related Work:} Addressing the issue of fake news requires a multidisciplinary effort, as the rise of misinformation in the internet age has highlighted the erosion of long-standing institutional bulwarks against it [...] \\
    \hline
    second\_least\_relevant & \textbf{Title:} Subsidy strategy of pharmaceutical e-commerce platform based on two-sided market theory \\ & \textbf{Abstract:} With the development of economic globalization and information technology, enterprises pay more attention to the sustainable development of their e-commerce. Under this trend, we study the subsidy strategy [...] \\ & \textbf{Related Work:} Previous research has extensively examined the subsidy strategy of pharmaceutical e-commerce platforms in two-sided markets. One paper, ``Competition in Two-Sided Markets," discusses the determinants [...] \\ 
    \hline
    least\_relevant & \textbf{Title:} Pedagogy in Cyberspace: The Dynamics of Online Discourse \\ & \textbf{Abstract:} This article elaborates a model for understanding pedagogy in online educational forums. The model identifies four key components. Intellectual engagement describes the foreground cognitive processes [...] \\ & \textbf{Related Work:} The concept of thought and language being distinct from each other is explored in one of the papers. The author argues that while there is a permanent interaction between thought and language [...] \\ \bottomrule
  \end{tabular}
  \caption{Example of each relevance category of an instance from the dataset.\label{dataset-example}}
\end{table}

\newpage
\section{Scikit-learn Models Hyperparameter Tuning}
\label{appendix-hyperparameter-tuning}
This appendix provides details on the hyperparameter tuning process for the four classifiers used in our experiments:  K-Nearest Neighbors (KNN), Random Forest (RF), Support Vector Classifier (SVC), and SGDClassifier\footnote{Linear classifiers (SVM, logistic regression, etc.) with SGD training. More info in https://scikit-learn.org/stable/modules/generated/sklearn.linear\_model.SGDClassifier.html} (SGD). We employed RandomizedSearchCV\footnote{https://scikit-learn.org/stable/modules/generated/sklearn.model\_selection.RandomizedSearchCV.html} for tuning, with three-fold cross-validation, 30 iterations, and PCA components [2, 50, 100, 200, None] implemented inside a scikit-learn pipeline\footnote{https://scikit-learn.org/stable/modules/generated/sklearn.pipeline.Pipeline.html}, along with specific hyper-parameter grids for each classifier, as outlined below.

\begin{table*}[ht]
\caption{Model Performance with Best Hyperparameters}\label{table-ml-models-performance}
\begin{tabular}{p{2cm}p{12cm}p{2cm}}
 \toprule
 Models          & Best Hyperparameters & Best score \\ \midrule
 KNN  &  \textbf{weights:} distance, \textbf{n\_neighbors:} 16, \textbf{metric:} euclidean, \textbf{algorithm:} auto, \textbf{pca\_ncomponents:} 50  & 0.719     \\
RF    &  \textbf{n\_estimators:} 500, \textbf{min\_samples\_split:} 10, \textbf{min\_samples\_leaf:} 6, \textbf{max\_features:} sqrt, \textbf{max\_depth:} 20, \textbf{pca\_ncomponents:} 200                   &            0.740 \\
 SVC            & \textbf{kernel:} rbf, \textbf{gamma:} scale, \textbf{C:} 1.0, \textbf{pca\_ncomponents:} 100                      &       \textbf{0.755}      \\
 SGD            &  \textbf{alpha:} 0.0365, \textbf{loss:} log, \textbf{max\_iter:} 184, \textbf{penalty:} l2, \textbf{pca\_ncomponents:} None               &         0.737    \\ \bottomrule
\end{tabular}
\end{table*}

\subsection{KNeighborsClassifier Hyperparameters}
\begin{verbatim}
`n_neighbors': np.arange(1, 21)
`weights': [`uniform', `distance']
`algorithm': [`auto', `ball_tree', `kd_tree', `brute']
`metric': [`euclidean', `manhattan', `minkowski']
\end{verbatim}

\subsection{RandomForestClassifier Hyperparameters}
\begin{verbatim}
`n_estimators': np.arange(50, 501, 50)
`max_depth': [None] + list(np.arange(10, 101, 10))
`min_samples_split': np.arange(2, 11)
`min_samples_leaf': np.arange(1, 11)
`max_features': [`auto', `sqrt', `log2'] 
\end{verbatim}

\subsection{SVC Hyperparameters}
\begin{verbatim}
`C':np.logspace(-3, 2, 6)
`gamma':[`scale', `auto']
`kernel': [`linear', `poly', `rbf', `sigmoid']
\end{verbatim}

\subsection{SGDClassifier Hyperparameters}
\begin{verbatim}
`loss': [`hinge', `log', `modified_huber']
`alpha': uniform(0.0001, 0.1)
`penalty': [`l2', `l1', `elasticnet']
`max_iter': randint(100, 1000)
\end{verbatim}

\subsection{Hyperparamater Tunning Results}
\label{subsec: Hyperparamater Tunning Results}

\autoref{table-ml-models-performance} provides the best hyperparameters for each model received after RandomizedSearchCV with the Kendall's Tau score on the test set using the best\_model\_ from RandomizedSearchCV.

\newpage
\clearpage

\section{SVC Results}
\label{appendix-f1-train-sizes}
In this section, we evaluate the data efficiency of our optimal sklearn classifier, SVC, by analyzing its performance across various classes using the F1-score. We methodically increase the training set size from 78 to $80\,524$ and assess performance on the test set of  $20\,132$ examples, with error bars derived via bootstrapping the test set 1000 times. We observe a performance imbalance across classes: there is a performance plateau for the `most relevant' class, which is of primary interest, along with the `second most relevant' and `least relevant' classes. Notably, increasing the training set size does not significantly improve performance for these classes. Furthermore, the `second least relevant' class proves to be the most challenging to predict. Although increasing the training set size does yield performance improvements for this class, its final performance remains substantially lower than that of other groups.
We trained this SVC model on the Kaggle platform, which took approximately $19.41$ minutes to train on the full train set ($80\,524$ instances).

\begin{figure}[H]
  \centering
  \includegraphics[width=0.8\textwidth]{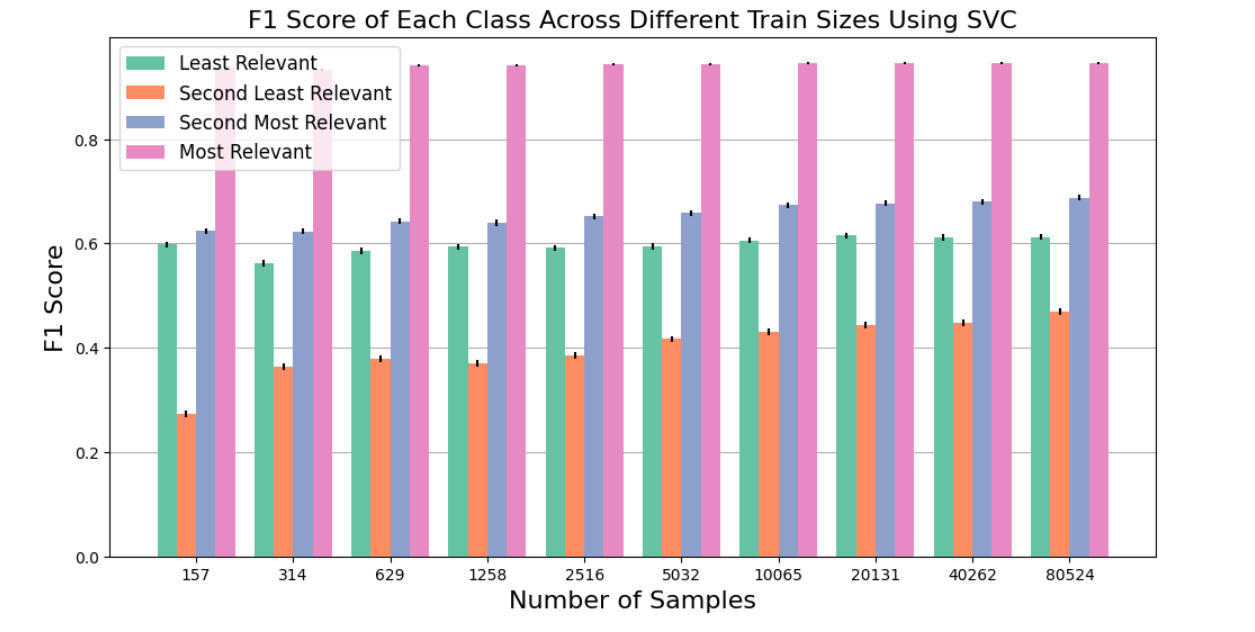}
  \caption{SVC Performance with Different Training Sizes and Corresponding F1-scores.\label{fig:classifier_with_train_sizes}}
\end{figure}

\newpage
\section{BERT Results}
\label{appendix-Bert-Results}

In sections \ref{appendix-bert-learning-curves} and \ref{appendix-f1-train-sizes-Bert}, we evaluate the data efficiency of BERT models using One-hot and Thermometer encoding, and compare them with SVC (in the case of Kendall's Tau score) by analyzing its performance across various training sets. We methodically increase the training set size from 157 to $80\,524$ (on a logarithmic scale) and assess performance on the test set of $20\,132$ examples, with error bars derived via bootstrapping the test set. The test set is the same for all experiments.

Section \ref{appendix-bootstrap-train-Bert} presents the result of the experiment for the hypothesis proposed in Section \ref{BERT}

\subsection{Performance of BERT Across Different Training Sets}
\label{appendix-bert-learning-curves}
\begin{figure}[H]
  \centering
  \includegraphics[width=0.75\textwidth]{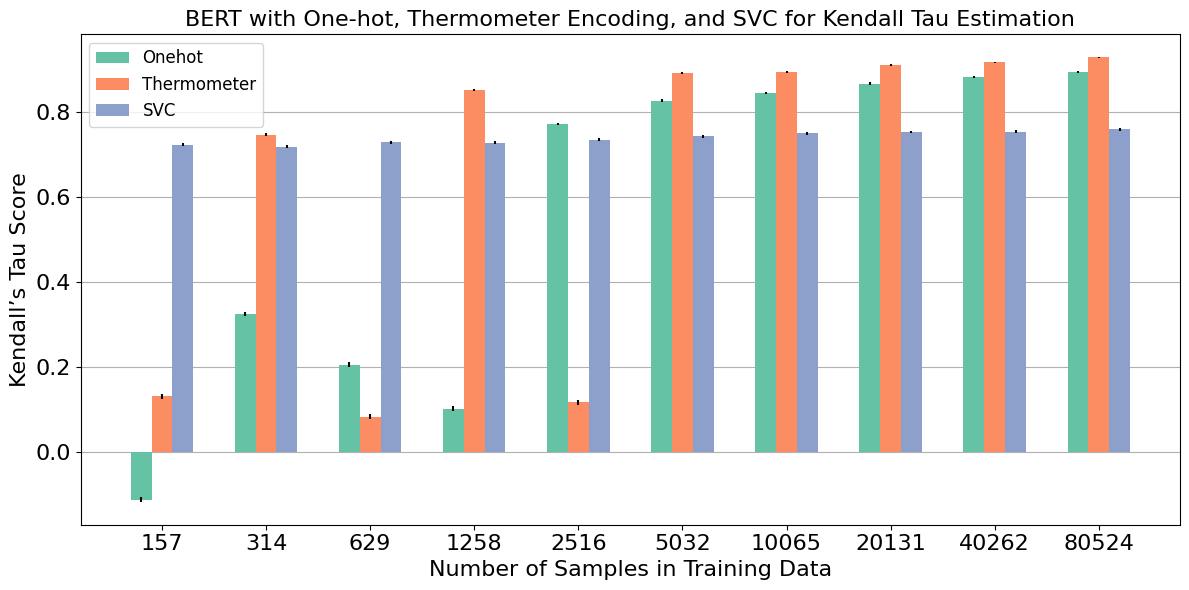}
  \caption{Kendall's Tau: BERT One-hot, Thermometer, SVC with Varying Data Sizes\label{fig:thermometer_vs_onehot_vs_svc_comparision}}
\end{figure}

We observe a fluctuation of performance in BERT models during the first 5 training sets, while in SVC the performance remains stable. When the training size is big enough (more than $5\,032$ examples), all 3 models do not yield significant performance improvements with the increase of the training set size.

\subsection{F1-score Comparison Across Different Training Sets}
\label{appendix-f1-train-sizes-Bert}

In terms of the F1-score across four classes, both BERT models using One-hot and Thermometer encoding show the same pattern, with very high scores in most and second most relevant classes, while encountering challenges in accurately classifying the least and second least relevant ones. This could be attributed to the dataset's construction, where the most relevant paper is the original one and the second most relevant is a paper cited by the original paper. As a result, the model finds it easier to identify these categories, leading to higher performance in those classes.

\begin{figure}[H]
  \centering
  \includegraphics[width=0.8\textwidth]{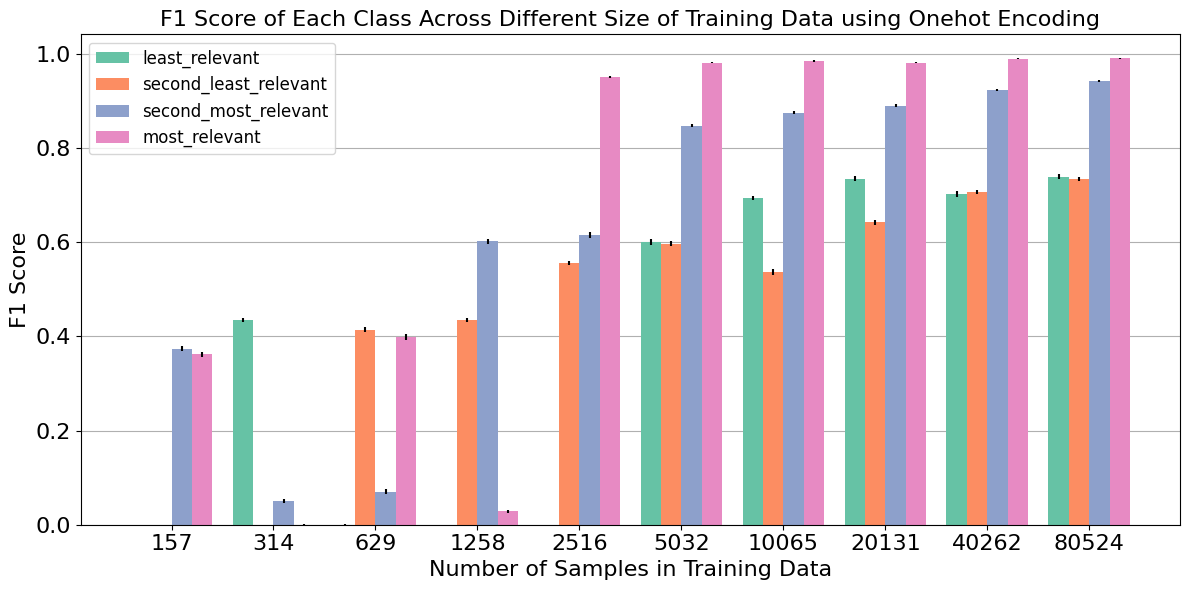}
  \caption{F1-score: BERT with One-hot encoded labels\label{fig:f1_onhot}}
\end{figure}

\begin{figure}[H]
  \centering
  \includegraphics[width=0.8\textwidth]{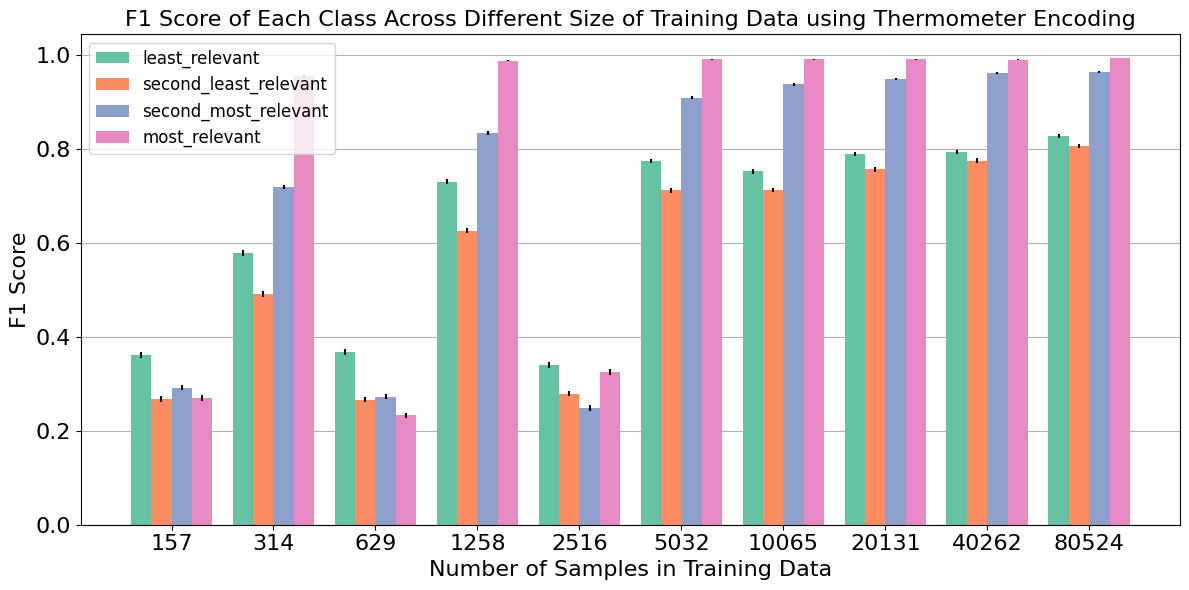}
  \caption{F1-score: BERT with Thermometer encoded labels\label{fig:f1_therm}}
\end{figure}

\subsection{Bootstrapped Training Sets}
\label{appendix-bootstrap-train-Bert}
 In the previous section, it is noticeable that the performances of BERT\textsubscript{onehot} and BERT\textsubscript{thermometer} fluctuate across the first training sets sizes when the number of samples is still small. This variability could potentially be attributed to the randomness inherent in the training data. Hence, we conducted an additional experiment to validate this hypothesis.
In this iteration, we applied bootstrapping to the training data 10 times for each subset within the initial 6 training sizes from 157 to $5\,032$ on a logarithmic scale. Subsequently, we made predictions on a fixed test dataset to compute the mean and standard error of Kendall’s Tau score.

\begin{figure}[H]
  \centering
  \includegraphics[width=0.8\textwidth]{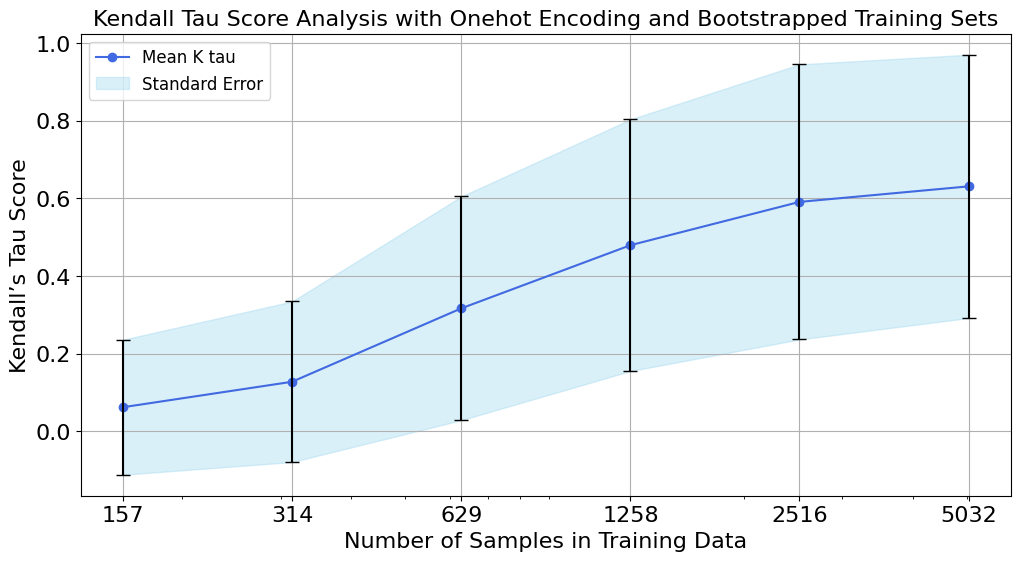}
  \caption{Kendall's Tau analysis: One-hot encoding\label{fig:tau_onhot_boot_train}}
\end{figure}

The learning curve of One-hot encoding shows increasing Kendall's Tau scores with larger training sizes, indicating better predictive performance. However, the expanding standard error suggests growing uncertainty despite fixed test data, likely due to variability introduced by bootstrapping the training data.

Observing the learning curve of Thermometer encoding, we notice a fluctuating trend in Kendall's Tau scores. Initially, there's a consistent increase, followed by a decrease, then subsequent increases and stabilization. This fluctuation suggests varying model performance as training sizes change. Despite fluctuations in Kendall's Tau scores, the standard error is noticeably smaller than in previous iterations. This suggests improved stability in the model's predictions, potentially indicating enhanced robustness or reduced variability in the bootstrapped training samples.

\begin{figure}[H]
  \centering
  \includegraphics[width=0.8\textwidth]{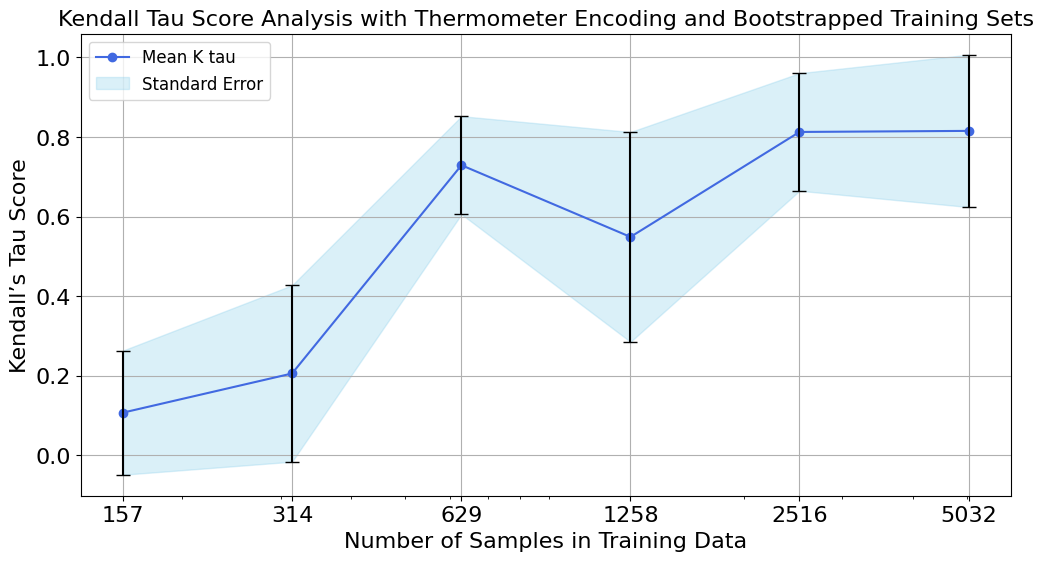}
  \caption{Kendall's Tau analysis: Thermometer encoding\label{fig:tau_therm_boot_train}}
\end{figure}

\newpage
\section{Cosine Similarity Baseline -- Thresholds definition algorithm}
\label{apx-cosine-thresholds}
The simplest baseline approach we have used to serve as reference for the other models was a classification based on fixed thresholds for the cosine similarity between the vector representations of prompts and texts. In order to choose the optimal values for said thresholds, we used the train set to find which combinations provided the best performance, using the Kendall's Tau as metric. To achieve this, we have created a simple algorithm made of three nested loops, one for each threshold, so that we could define a grid with all the combinations for the three thresholds values, ranging from the minimum similarity value to the maximum one, with a step of 0.025.

For each iteration, the instances of the Train set were classified according to the current thresholds:

\begin{itemize}
  \item \textbf{least\_relevant: } Similarity smaller than the first threshold;
  \item \textbf{second\_least\_relevant: } Similarity between first and second thresholds;
  \item \textbf{second\_most\_relevant: } Similarity between second and third thresholds;
  \item \textbf{most\_relevant: } Similarity greater than the third threshold;
\end{itemize}

\begin{figure}[H]
  \centering
  \includegraphics[width=0.9\textwidth]{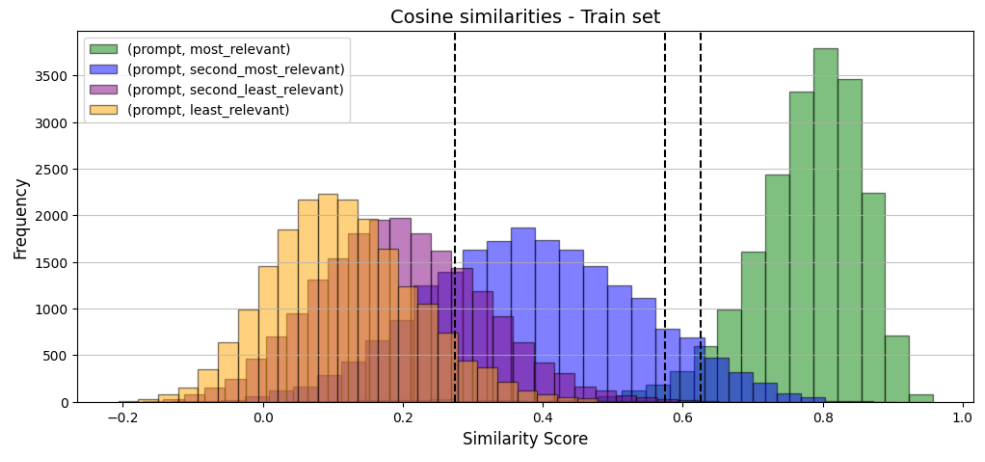}
  \caption{Optimal thresholds found in train set using Kendall's Tau as metric\label{fig:cos_thresholds_train}}
\end{figure}

Finally, we computed the Kendall's Tau in the Train set for every combination of the grid, comparing the real labels with the classifications made according to the thresholds. The optimal thresholds values found were [0.275, 0.575, 0.625], achieving the highest overall Kendall's Tau coefficient of 0.775 on the Train set. The \autoref{fig:cos_thresholds_train} visually displays the three thresholds (dashed lines) in the train set distribution per relevance category.

\begin{figure}[H]
  \centering
  \includegraphics[width=0.9\textwidth]{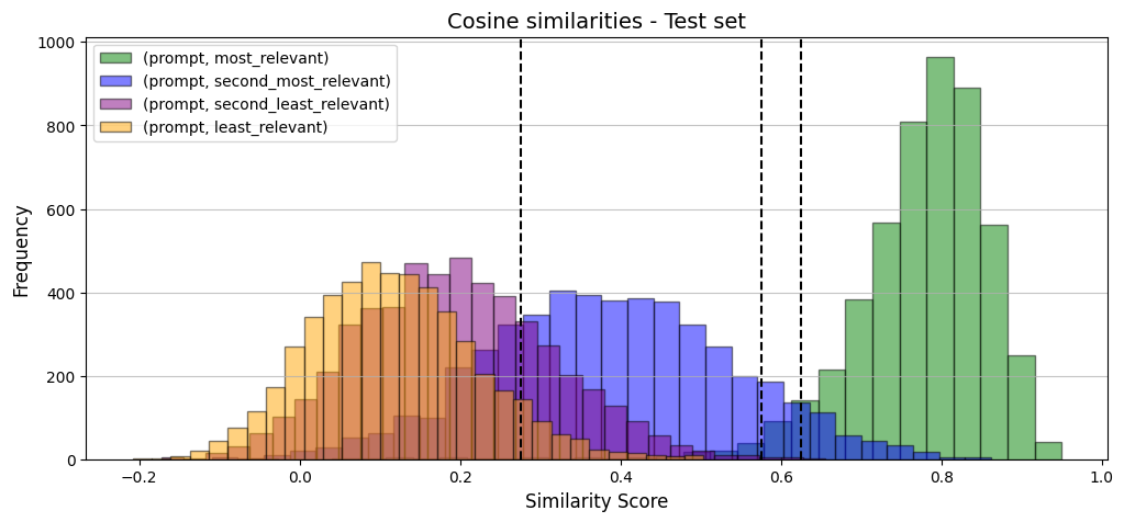}
  \caption{Optimal thresholds found in test set using Kendall's Tau as metric\label{fig:cos_thresholds_test}}
\end{figure}

Once the thresholds had been determined in the train set, we measured this approach's performance in the test set as well in order to compare it with the different methodologies studied in this paper. \autoref{fig:cos_thresholds_test} below visually displays the three thresholds (dashed lines) in the test set distribution per relevance category, where the model obtained a Kendall's Tau of 0.774. Visually comparing both distribution plots, we notice that the train set and test set are indeed very similar, which may explain why the Kendall's Tau for the test data was almost identical to the one obtained for the train data. The performance of this baseline was further studied through bootstrapped samples of the test set in order to obtain confidence intervals for the analysed metric. These results are displayed on \autoref{tab:performance} in \autoref{comparison-performances}.

\newpage
\section{Scrapped subjects across 22 different fields}
\label{subject-scrapped}
\begin{multicols}{3}
\begin{itemize}
  \item Computer Science
  \item Medicine
  \item Chemistry
  \item Biology
  \item Materials Science
  \item Physics
  \item Geology
  \item Psychology
  \item Art
  \item History
  \item Geography
  \item Sociology
  \item Business
  \item Political Science
  \item Economics
  \item Engineering
  \item Education
  \item Env Science
  \item Philosophy
  \item Mathematics
  \item Law
  \item Linguistics
\end{itemize}
\end{multicols}

\end{document}